\documentclass{article}

\usepackage{times}
\usepackage{graphicx} 
\usepackage{subfigure} 

\usepackage{natbib}

\usepackage{hyperref}

\usepackage[accepted]{icml2018}

\usepackage{makecell}
\usepackage{graphicx}
\usepackage{amsmath}
\usepackage{amsthm}
\usepackage{amssymb}
\usepackage{bbm}
\usepackage{algorithm}
\usepackage{algorithmic}
\usepackage{color}
\usepackage{booktabs}
\usepackage{multirow}

\usepackage{symbols}

\makeatother

\newcommand*{\ShowNotes}{}
\definecolor{darkred}{rgb}{0.7,0.1,0.1}
\definecolor{darkgreen}{rgb}{0.1,0.7,0.1}
\definecolor{cyan}{rgb}{0.7,0.0,0.7}
\definecolor{dblue}{rgb}{0.2,0.2,0.8}
\definecolor{maroon}{rgb}{0.76,.13,.28}
\definecolor{burntorange}{rgb}{0.81,.33,0}
\definecolor{tealblue}{rgb}{0.212,0.459, 0.533}

\ifdefined\ShowNotes
  \newcommand{\colornote}[3]{{\color{#1}\bf{#2: #3}\normalfont}}
\else
  \newcommand{\colornote}[3]{}
\fi

\icmltitlerunning{Adversarial Posterior Distillation}

\begin{document} 

\twocolumn[
\icmltitle{Adversarial Distillation of Bayesian Neural Network Posteriors}



\icmlsetsymbol{equal}{*}
\begin{icmlauthorlist}



\icmlauthor{Kuan-Chieh Wang}{to,vi}   
\icmlauthor{Paul Vicol}{to,vi}
\icmlauthor{James Lucas}{to,vi}
\icmlauthor{Li Gu}{to}
\icmlauthor{Roger Grosse}{to,vi}
\icmlauthor{Richard Zemel}{to,vi}

\end{icmlauthorlist}
\icmlaffiliation{to}{University of Toronto, Toronto, Ontario, Canada}
\icmlaffiliation{vi}{Vector Institute, Toronto, Ontario, Canada}
\icmlcorrespondingauthor{Kuan-Chieh Wang}{wangkua1@cs.toronto.edu}
\icmlkeywords{generative adversarial nets, bayesian neural networks, amortized inference}

\vskip 0.3in
]



\printAffiliationsAndNotice{}  


\begin{abstract}

Bayesian neural networks (BNNs) allow us to reason about uncertainty in a principled way.
Stochastic Gradient Langevin Dynamics (SGLD) enables efficient BNN learning by drawing samples from the BNN posterior using mini-batches.
However, SGLD and its extensions require storage of many copies of the model parameters, a potentially prohibitive cost, especially for large neural networks.
We propose a framework, Adversarial Posterior Distillation, to distill the SGLD samples using a Generative Adversarial Network (GAN).
At test-time, samples are generated by the GAN. We show that this distillation framework incurs no loss in performance on recent BNN applications including anomaly detection, active learning, and defense against adversarial attacks.
By construction, our framework distills not only the Bayesian predictive distribution, but the posterior itself.
This allows one to compute quantities such as the \textit{approximate model variance}, which is useful in downstream tasks.
To our knowledge, these are the first results applying MCMC-based BNNs to the aforementioned applications.

\end{abstract}


\vspace{-0.2cm}
\section{Introduction}\label{sec_intro}
\vspace{-0.1cm}

Neural networks (NNs) are often viewed as powerful black-box systems whose behaviors are difficult to interpret and control. Despite significant progress made on supervised learning with Stochastic Gradient Descent (SGD), users are still apprehensive when applying NNs in safety-critical systems. Moreover, recent work has shown that modern neural networks can be miscalibrated and may be over-confident about their predictions \cite{guo2017calibration}.
The overarching motivation of this research is to have reliable estimates of uncertainty when making predictions with a NN.

Uncertainty is important in many scenarios.
For example, designers of autonomous cars might want passengers to take control when the system is uncertain about the scene.
Uncertainty can also be used to understand a system, such as in the prediction of basketball player movements; a player's offensive skill can be gauged by the amount of uncertainty he is able to induce by his movements.

Bayesian methods provide a principled way to model uncertainty through the posterior distribution over model parameters. 
Most approaches for learning Bayesian Neural Networks (BNNs) \cite{mackay1992practical} fall into one of two categories: variational inference (VI) or Markov chain Monte-Carlo (MCMC).
The disadvantage of MCMC methods is their computational cost, both in terms of time and storage, and the difficulty in evaluation.
However, MCMC methods are appealing because, in the limit, they produce samples from the true posterior.
In contrast, VI methods require one to choose a family of approximating distributions, and thus can only produce samples from an approximate posterior.

The main goal of our work is to reduce the storage overhead involved in maintaining MCMC samples, and to show the usefulness of MCMC methods in modern BNN applications.
We employ Generative Adversarial Networks (GANs) \cite{goodfellow2014generative} to model the MCMC samples.
This yields a parametric approximation (i.e., the generator) of the distribution of MCMC samples, that eliminates the storage overhead while providing access to the posterior at test-time.
We evaluate our approach on a range of applications including classification, anomaly detection, active learning, and defense against adversarial examples, and show that the distilled samples perform as well as the original MCMC samples.
We also analyze the suitability of GANs for this distillation process, taking into account recent advances in stabilizing GANs.


\vspace{-0.2cm}
\section{Background}\label{sec_back}
\vspace{-0.1cm}

In this section, we provide a brief overview of BNNs and the technical background required for our study.

\subsection{Uncertainty Estimates with BNNs}

In the standard deterministic NN setup, we aim to optimize the network parameters $\theta$ given a loss function $\mathcal{L}$ and a dataset of input-output pairs $\mathcal{D} = \{(x_i, y_i)\}_{i=1}^N$ as follows:
\begin{equation}
\theta = \text{argmin}_{\theta'}\mathcal{L}(\theta', \mathcal{D})
\end{equation}
where typically the loss is the regularized negative log-likelihood:
\begin{equation}
\label{eq:nll}
\mathcal{L} = -\sum_i \log p(y_i | x_i,\theta') + \norm{\theta'}_2^2
\end{equation}
This is equivalent to MAP estimation of a BNN with a Gaussian prior, $\theta \sim \mathcal{N}(0,I) = p(\theta)$.
At test time, one uses the learned parameters to make predictions.

In contrast to a deterministic NN, where parameters are represented by a point estimate, the parameters in a BNN are represented by probability distributions.
Given a prior distribution $p(\theta)$ over model parameters, the goal is to obtain the posterior $p(\theta | \mathcal{D})$.
At test time, instead of using the point-estimate approximation, one needs to marginalize out the posterior.
We provide a more detailed discussion in Section~\ref{sec_app}.

\subsection{Stochastic Gradient Langevin Dynamics} 
Stochastic Gradient Langevin Dynamics (SGLD) is a pivotal work for the application of MCMC methods to BNNs ~\cite{welling2011bayesian}.  Each iteration of MCMC traditionally requires computation over the full dataset (e.g., to compute the Metropolis-Hastings acceptance ratio).
\citet{welling2011bayesian} develop theoretical justification for learning the posterior using mini-batches of data.
Given a standard classification problem with a loss function as given in Eqn.~\ref{eq:nll}, the usual SGD update can be written as:
\begin{equation}
\label{eq:sgd}
\Delta \theta^t = \frac{\epsilon^t}{2} \left( \nabla \log p(\theta^t) + \frac{N}{n} \sum_{i=1}^n \nabla \log p(y_i^t | x_{i}^t, \theta^t) \right)
\end{equation}
where $n$ is the mini-batch size, and superscript $t$ denotes the update iteration. Then, the SGLD update is just:
\begin{equation}
\label{eq:sgld}
\Delta \theta^t = \frac{\epsilon^t}{2} \left( \nabla \log p(\theta^t) + \frac{N}{n} \sum_{i=1}^n \nabla \log p(y_i^t | x_{i}^t, \theta^t) \right) + \eta^t
\end{equation}
where $\eta^t \sim \mathcal{N}(0, \epsilon^t)$. Note that Eqn. \ref{eq:sgld} is simply Eqn. \ref{eq:sgd} with added Gaussian noise.
\citet{welling2011bayesian} show that when the step-size $\epsilon^t$ is decayed polynomially, the process transitions from stochastic optimization to Langevin dynamics, where each update yields an unbiased sample from the posterior.

In this work, we use SGLD to obtain samples of $\theta$.
However, our framework is also compatible with other extensions to SGLD; these extensions are complementary to our method.

\subsection{Generative Adversarial Networks}
\label{sec:gan}

Generative Adversarial Networks (GANs) are an approach to generative modeling that consist of two components: a generator, $G$, maps random noise $z \sim \mathcal{N}(0, I)$ to approximate data samples $G(z)$; and a discriminator, $D$, tries to distinguish between generated data $G(z)$ and real data $x$.
$G$ is trained to confuse $D$.
In our approach, the ``data'' are the model samples $\theta$, hence we write $\theta$ instead of $x$ when appropriate.
The GAN objective is:
\vspace{-2mm}
\begin{equation}
\min_G \max_D \mathbb{E}_{\theta \sim p(\theta|\mathcal{D})} [\log D(\theta)] + \mathbb{E}_{z \sim p_z(z)} [\log (1 - D(G(z)))]
\end{equation}
A recent GAN formulation with empirical success uses Wasserstein distance as the loss \cite{arjovsky2017wasserstein}.
In \cite{arjovsky2017wasserstein}, $D$ is required to be Lipschitz continuous, and weight clipping is used as a crude approximation.
Extensions include using a gradient penalty \cite{gulrajani2017improved}, and finite differences of gradients \cite{anonymous2018improving} to enforce this constraint.  Other formulations of the Wasserstein distance have also been studied \cite{anonymous2018improvingOT}.
In this work, we opt to use the WGAN with gradient penalty (WGAN-GP)~\cite{gulrajani2017improved}, which optimizes the following objective:
\begin{equation}
\begin{aligned}
\mathcal{L} &= \mathbb{E}_{\tilde{\theta} \sim P_g} [D(\tilde{\theta})] - \mathbb{E}_{\theta \sim P_r} [D(\theta)] \\ &+ \lambda \mathbb{E}_{\hat{\theta} \sim P_{\hat{\theta}}} [(|| \nabla_{\hat{\theta}} D(\hat{\theta}) ||_2 - 1)^2]
\end{aligned}
\end{equation}

\paragraph{Evaluation.} Despite recent advances in GANs, evaluation remains a challenge.
Most research in GANs deals with image generation, for which it is difficult to quantify performance.
Some metrics are based on visual quality, including the InceptionScore \cite{impgan} and unsupervised SSIM \cite{gan-e-ssim}.
Similarly, evaluating the quality of MCMC posterior samples has long been a challenge.
In this work, because our samples are network parameters, we have the opportunity to quantitatively evaluate the samples---both the original SGLD samples and the ones generated by the GAN---by applying the BNN with those parameters across a range of applications (see Section~\ref{sec_relate_work}).


\vspace{-0.2cm}
\section{Method}
\label{sec_app}
\vspace{-0.1cm}

\begin{figure}[t]
    \centering
    \includegraphics[width=\linewidth]{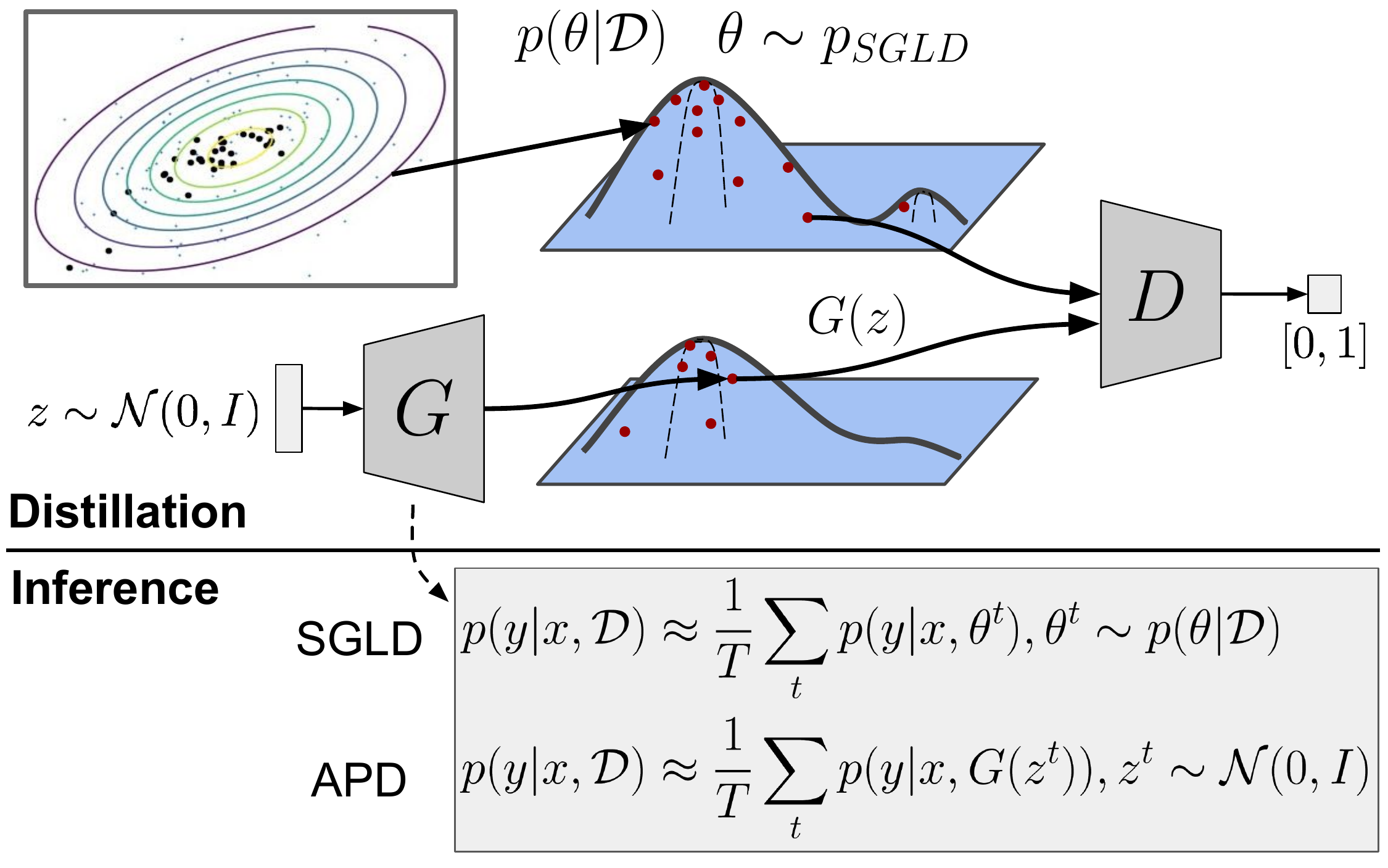}
    \caption{\textbf{APD Framework.} \textit{Distillation:} Posterior samples are generated from the target network (top) and used to train the generator network (bottom). \textit{Inference:} When performing inference, we sample from the generator network.}
    \label{fig:adversarialsgld}
\end{figure}

In this section, we introduce a framework for Bayesian inference that consists of two steps: 1) obtain a set of samples from the posterior distribution over network parameters $\theta \sim p(\theta | \mathcal{D})$ using SGLD; 2) train a WGAN-GP to model the posterior samples. This process is illustrated in  Figure~\ref{fig:adversarialsgld}. The result is a single generative model that distills the posterior distribution; this allows us to draw samples efficiently (i.e., in parallel, as opposed to traditional MCMC steps which are performed sequentially), with little storage overhead (i.e., we only need to store the parameters of a relatively small generator).
We call this formulation Adversarial Posterior Distillation (APD).

\begin{algorithm}[t]
\caption{Offline APD}\label{algo:acb}
\begin{algorithmic}[1]
\STATE Sample $\{\theta^t\}_{t=1}^T$ using MCMC updates, where $T$ denotes the number of updates.
\STATE Optimize $G$ with WGAN-GP loss using $\{\theta^t\}_{t=1}^T$ as real data. 
\end{algorithmic}
\end{algorithm}

The distillation process can be performed either \textit{offline} or \textit{online}.
We outline the offline variant of APD in Algorithm \ref{algo:acb}.
This requires the user to store a large number of samples prior to distillation.
We further generalize this framework to the online setting where MCMC steps and GAN updates are interleaved (Algorithm \ref{algo:acbi}). Both variants perform similarly when the buffer $\theta_R$ is reasonably large. 

\begin{algorithm}[t]
\caption{Online APD }\label{algo:acbi}
\begin{algorithmic}[1]
\STATE Initialize $\{\theta^{0,k}\}_{k=1}^K$, the $K$ independent BNN parameters using generated samples from $G$. 
\WHILE{not converged}
\STATE Sample $\{\theta^{t,k}\}_{t=t}^{t+T_m}$ using MCMC updates, where $T_m$ denotes the number of updates for all $k$.
\STATE Add $\{\theta^{t,k}\}$ to $\theta_R$ denoting a buffer of samples to distill
\STATE Optimize $G$ with WGAN-GP loss using $\theta_R$ as real data for $T_g$ gradient steps. 
\ENDWHILE
\end{algorithmic}
\end{algorithm}
\vspace{-3mm}

\paragraph{Details.} For SGLD, instead of using the sampling scheme suggested by \citet{welling2011bayesian}, we follow \citet{balan2015bayesian}, where the number of burn-in iterations and the sampling interval are treated as hyperparameters instead of monitoring when SGLD transitions into the sampling phase.
Similarly to \citet{balan2015bayesian}, we use a fixed learning rate, as we found that this led to better performance in our experiments. It has been shown that the bias introduced by this modified version of SGLD is quantifiable \citep{vollmer2016exploration}.

\subsection{Uncertainty Estimates}
Once the generator is sufficiently trained, we can discard the SGLD posterior samples and use the generator to produce samples at test time, for use in downstream tasks.
The common feature of the downstream applications we use for evaluation is that they require an estimate of uncertainty.  Here we outline the uncertainty estimates used in Section~\ref{sec_exp}.

For prediction using APD, we perform MC integration by drawing samples from the trained generator:
\begin{align}\label{eqn:mc_integration}
p(y |x,\mathcal{D}) &= \mathbb{E}_{\theta | \mathcal{D}}[p(y | x,\theta)] \\
&\approx \frac{1}{T}\sum_{t=1}^T p(y | x, G(z^t)), \quad z^t \sim \mathcal{N}(0,I)
\end{align}
To measure uncertainty, we can compute the \textbf{entropy}, $\text{H}(y|x, \mathcal{D})$
or the Bayesian Active Learning by Disagreement objective (\textbf{BALD}) \citep{houlsby2011bayesian}:
\begin{equation}
\mathbb{I}(y,\theta|x,\mathcal{D}) = \text{H}(y|x, \mathcal{D}) - \mathbb{E}_{\theta|\mathcal{D}}[\text{H}(y|x,\theta)]
\end{equation}
or the \textbf{variations-ratio} (VR):
\begin{align}
\begin{split}
\text{VR}(x) &= 1 - \frac{1}{T}\sum_{t} \mathbbm{1}[y^t=c^*], \\
c^* &= \text{argmax}_c \sum_t \mathbbm{1}[y^t=c], \text{ where } c \text{ indexes classes}\\
\quad y^t &= \text{argmax}_y p(y | x, G(z^t))
\end{split}\label{eqn:var_ratio}
\end{align}
or the \textbf{approximate model variance} as defined in \citet{feinman2017detecting}:
\begin{equation}
    U(x) = \dfrac{1}{T} \sum_{t=1}^{T}\bp_t^T \bp_t - ( \dfrac{1}{T} \sum_{t=1}^{T}\bp_t)^T ( \dfrac{1}{T} \sum_{t=1}^{T}\bp_t)
\label{eqn:uncert}
\end{equation}
where $\bp_t$ are the stochastic vectors of class predictions.


\vspace{-0.2cm}
\section{Related Work}\label{sec_relate_work}
\vspace{-0.1cm}

In this section, we give an overview of recent BNN learning methods and modern applications of BNNs, to motivate our experimental studies.

\subsection{Learning BNNs}

\paragraph{Variational Inference (VI).}
VI methods construct an approximating distribution $q(\theta) \approx p(\theta|\mathcal{D})$ and optimize to make the approximation close to the true posterior.
This often involves making assumptions such as that the parameters can be fully-factorized, as $p(\theta \mid \mathcal{D}) = \prod p(w_i \mid \mathcal{D})$.

VI was first proposed for neural networks by \citet{hinton1993keeping}. \citet{graves2011practical} made VI practical by introducing a stochastic VI method with a diagonal Gaussian posterior. Graves' method uses a biased Monte Carlo estimate of the variational lower bound. Later, \citet{blundell2015weight} introduced an algorithm for training BNNs called Bayes by Backprop (BBB), that uses an unbiased Monte Carlo estimate of the lower bound, based on the reparameterization trick \cite{kingma2014stochastic, rezende2014stochastic}.

An alternative approach is Expectation Propagation (EP) \cite{minka2001expectation}, a deterministic approximation method that extends \textit{assumed density filtering (ADF)} by iteratively refining the approximations. Probabilistic Backpropagation (PBP) \cite{hernandez2015probabilistic} is a recently-introduced online extension of EP. 

More recently, \citet{louizos2017multiplicative} proposed to use the idea of a \textit{flow} to break the simplifying mean-field assumption, but this method is still restricted by the kinds of transformations allowed by a flow (e.g., invertible).  Bayesian Hypernetworks (BH) \cite{krueger2017bayesian} use a hypernetwork to generate shift and scale distribution over network activations, where the hypernet is restricted to be an invertible generative model.

\paragraph{MCMC}
MCMC methods have long been used to learn BNNs \cite{neal1996bayesian}. However, traditional MCMC methods require computation over the whole dataset per iteration.
Since the introduction of SGLD, a suite of stochastic-gradient MCMC algorithms have been proposed \cite{ahn2012bayesian,ahn2014distributed,balan2015bayesian,chen2014stochastic,ding2014bayesian}, drawing from the wealth of knowledge behind general MCMC techniques. This demonstrates the potential of MCMC methods for BNNs. However, to make use of these methods, one needs to store a sufficient number of posterior samples, which incurs significant storage overhead.

\citet{balan2015bayesian} proposed a method to approximate the Bayesian predictive distribution using a single network. They train a student model $\mathcal{S}(y | x, w)$ to approximate the Bayesian predictive distribution $q(y | x)$ by minimizing the KL divergence $D_\text{KL} [ \mathbb{E}_{\theta|\mathcal{D}}[p(y|x,\theta)] || \mathcal{S}(y | x, w)]$. 
This avoids the storage cost and integration at test-time.  However, the posterior $p(\theta|\mathcal{D})$ is lost at test-time, which makes this method unfit for cases where the posterior is required for other computations.
In contrast, our formulation aims to distill the posterior to be sampled from for downstream use.

\citet{li2017approximate} studied a framework that most resembles ours. Both their work and ours use a GAN to replace MCMC samples, yet the setting and goals are different.
\citet{li2017approximate} only provided results related to BNNs on small NNs with 50 hidden units on binary classification accuracy/loss. They instead explored additional tasks such as using their sampler to improve latent variable inference for missing-data imputation.

\subsection{Recent BNN Applications}
A number of interesting applications of BNNs have been studied in the context of recent VI methods.
Similarly, MC dropout \cite{gal2016dropout}---which is a simple approximation that applies dropout \cite{srivastava2014dropout} at test-time---has recently been used in real-world applications.
Below we summarize these results by task.

\paragraph{Standard Classification/Regression.} Though perhaps not the most informative tasks for examining BNNs, regression loss or classification accuracy are usually reported by BNN studies. Some use 1-dimensional regression problems \cite{hernandez2015probabilistic}. Others use standard deep learning classification benchmarks such as MNIST \cite{mnist, balan2015bayesian,blundell2015weight,gal2016dropout}, or CIFAR10 \cite{krizhevsky2009learning, krueger2017bayesian, louizos2017multiplicative}, which we adopt as well.
Since good classification accuracy only requires a good point-estimate of $\theta$, additional tasks are usually used to evaluate BNNs.

\paragraph{Anomaly Detection.} Anomaly detection refers to detecting \textit{out-of-distribution} (OOD) data (such as white noise) given a BNN trained only on \textit{in-distribution} data (e.g., MNIST).  Intuitively, a good BNN should be more uncertain about OOD inputs, enabling better detection of OOD data.
\citet{hendrycks2016baseline} provide benchmarks for anomaly detection. \citet{krueger2017bayesian} applied BH to this task, and showed that both MC dropout and BH outperform deterministic NNs.
Hence, we provide detailed results using both existing MCMC-based methods and ours. 

\paragraph{Exploration.}
\citet{blundell2015weight} used Thompson sampling based on BNN outputs to minimize regret in a bandit problem. \citet{hernandez2015probabilistic} performed active learning on a 1-dimensional regression problem.  Both of these settings involve fairly small datasets and models (e.g., input size on the order of 10, and NN with 1-hidden layer of width 50). More recently, \citet{gal2017deep} achieved good results using MC dropout for active learning with image data (e.g., MNIST and real-world medical images).

\paragraph{Detecting Adversarial Attacks.} Adversarial examples \cite{szegedy2013intriguing} are inputs to a neural network that are designed to force misclassification.
These inputs often appear normal to humans but cause the neural network to make inaccurate predictions.
This raises an interesting question: can we train neural networks to detect adversarial examples? Bayesian neural networks are an obvious candidate for this task and thus they have been explored before \cite{feinman2017detecting, louizos2017multiplicative, rawat2017adversarial}. Related to our work, \citet{feinman2017detecting} used MC dropout inference to detect adversarial examples and showed promising results.
To the best of our knowledge, the use of SGLD for this task has not been explored.

Lastly, our work is focused on distilling BNNs and studying modern downstream applications. To the best of our knowledge, our work is the first to provide results for MCMC-based BNNs for these applications.


\section{Experiments}
\label{sec_exp}

In this section, we show that APD performs as well as the true SGLD samples in terms of classification accuracy, and on the recent anomaly detection benchmarks provided by \citet{hendrycks2016baseline}.
On other difficult tasks---active learning and defense against attacks---we show that SGLD does at least as well as MC dropout \cite{gal2016dropout}. APD did not match SGLD, but did perform as well as MC dropout in each case.
Lastly, studies show that distilling the posterior is a challenging task, and that recent advances in GANs improve APD. \footnote{Implementation details can be found at \url{https://github.com/wangkua1/apd_public}}

\subsection{Toy 2D Classification}

\begin{figure}[t]
    \centering
    \includegraphics[width=\linewidth]{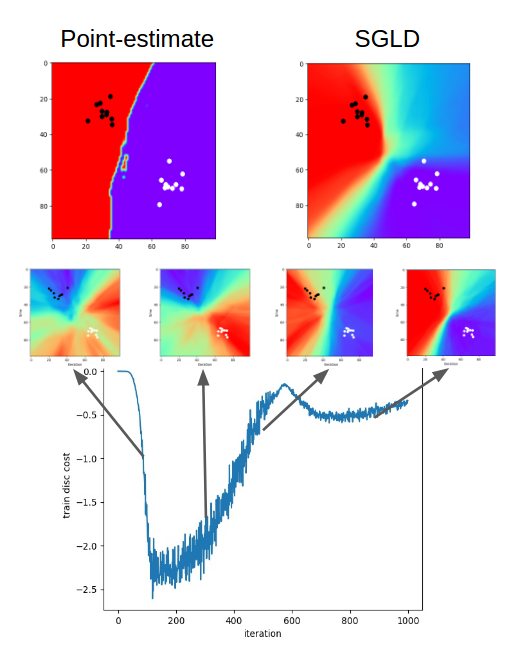}
    \vspace{-0.3cm}
    \caption
    {\textbf{Toy2D Classification Results.} \textit{Top}: The decision boundaries of models trained with SGD vs SGLD. The 10 $\times$ 2 white/black dots are inputs from each of the two classes. The model learned with SGD is very confident everywhere, whereas the one learned with SGLD is uncertain far from the training data.  \textit{Bottom}: The learning curve for the discriminator loss of WGAN-GP.  The initial generated samples result in a random decision boundary. \textit{Middle}: As the WGAN-GP loss improves, the decision boundary looks more similar to the one obtained with real SGLD samples.}
    \label{fig:toy2d}
    \vspace{-0.5cm}
\end{figure}

We first validated our approach using a 2D toy dataset following \citet{balan2015bayesian}. The dataset consists of two clusters of 10 points each in 2D space, easily separable by a linear classifier.  This toy task is used to determine the ability of a model to capture uncertainty far from the data distribution. 
We trained a simple NN to perform binary classification on this dataset, using both SGD and SGLD. We used a fully-connected NN (fcNN) with two hidden layers of 10 units each (2-10-10-2). The results are shown in Figure~\ref{fig:toy2d}. The predictive uncertainty increases in regions far from the observed data. APD was also able to capture this behavior.

\subsection{Predictive Performance and Uncertainty}

Recently, anomaly detection has been used as a benchmark
for BNNs. Here, we show that SGLD and APD are able
to detect anomalies better than the SGD and MC dropout
baselines.

\subsubsection{Experimental Setup}

We used MNIST for our classification and anomaly detection experiments.
We trained on 50,000 examples, and reserved 10,000 from the standard training set as a fixed validation set.
When training with SGD, we tuned the learning rate and weight decay on the validation set: we found the best values to be 0.05 and 0.001, respectively.
We did not use momentum, for fair comparison with vanilla SGLD, which did not use momentum.
For baselines, we used \textbf{SGD} and MC dropout (\textbf{MC-Drop}), where we used the same parameters as for SGD, with an additional dropout rate set to 0.5.
For SGLD, we did not use dropout, and the number of burn-in iterations and sampling interval were 500 and 20, respectively.
The batch size for training was fixed at 100 for all methods.
For the approximate BNNs---MC-Drop, SGLD, and APD---predictions were based on the MC estimate of 200 network samples unless otherwise specified.

We experimented with two fcNN architectures: \textbf{fcNN1}, with architecture 784-100-10 (79,510 parameters), and \textbf{fcNN2}, with architecture 784-400-400-10 (478,410 parameters).
For APD, we used a 3-layer fcNN with 100 hidden units per layer for both our generator and discriminator, for all tasks.

\subsubsection{Classification Accuracy}
We evaluated the classification accuracy of our method on MNIST, and compared to SGD, MC dropout, and SGLD.
The results are shown in Table~\ref{table:mnistclassification}. The architectures we explored use narrow hidden layers compared to typical dropout architectures but nonetheless contain a large number of parameters. We include these results to demonstrate that APD is able to distill the posterior distribution of large networks without sacrificing performance on this task.
Also, these networks, which performed reasonably on classification, were used for anomaly detection in the following subsection.

\begin{table}[!htbp]
\setlength{\tabcolsep}{4pt}
\footnotesize
\centering
\begin{tabular}{@{}cccccccc@{}}
\toprule
\textbf{Dataset} & \textbf{Model} & \textbf{SGD}    & \textbf{MC-Drop} & \textbf{SGLD}   & \textbf{APD (Ours)} \\ \midrule

\textbf{MNIST}   & \textbf{fcNN1} & 0.981  & 0.973   & 0.979  & 0.978   \\
\textbf{MNIST}   & \textbf{fcNN2} & 0.981  & 0.983   & 0.980  & 0.981   \\

\bottomrule
\end{tabular}
\caption{\textbf{MNIST Classification Results.} The samples of network parameters produced by APD achieve classification accuracy competitive with other BNN methods, including MC dropout and SGLD.}
\label{table:mnistclassification}
\end{table}

\vspace{-0.1cm}
\subsubsection{Anomaly Detection}
\label{sec:anomdet}

We measured the performance of our method on the MNIST anomaly detection task introduced by~\citet{hendrycks2016baseline} and used by \citet{louizos2017multiplicative} and \citet{krueger2017bayesian}.
Training was unmodified from the previous subsection.
At test-time, the inputs consisted of both \textit{in-distribution} and OOD data.
We evaluated performance using the area under receiver operating curve (AUROC) and the area under precision-recall curves (AUPR+/-). ROC is the curve of true-positive rate versus false-positive rate. AUPR+/- is similar, but adjusts for different base rates between the two classes.
For both metrics, higher numbers indicate better detection performance.
We refer readers to ~\citet{hendrycks2016baseline} for details. \footnote{We adapted the evaluation code provided at \url{http://github.com/hendrycks/error-detection}}

Although \citet{hendrycks2016baseline} and \citet{krueger2017bayesian} showed reasonable results, the deterministic baseline NN already performed very well on this task (i.e., $>90\%$).
We found that the baseline is susceptible to scaling of the pixel intensity of the OOD data (i.e., when the intensities are multiplied by a scalar value $\neq1$). Hence, for our experiments we scaled the OOD datasets by a factor of 5.
Table~\ref{mnistanomalydetection} shows that our method outperforms SGD and MC dropout, and is competitive with SGLD.
An analysis of the effect of OOD scaling and more anomaly detection results are provided in the appendix.

\begin{figure}[htbp]
    \centering
    \includegraphics[width=
    \linewidth]{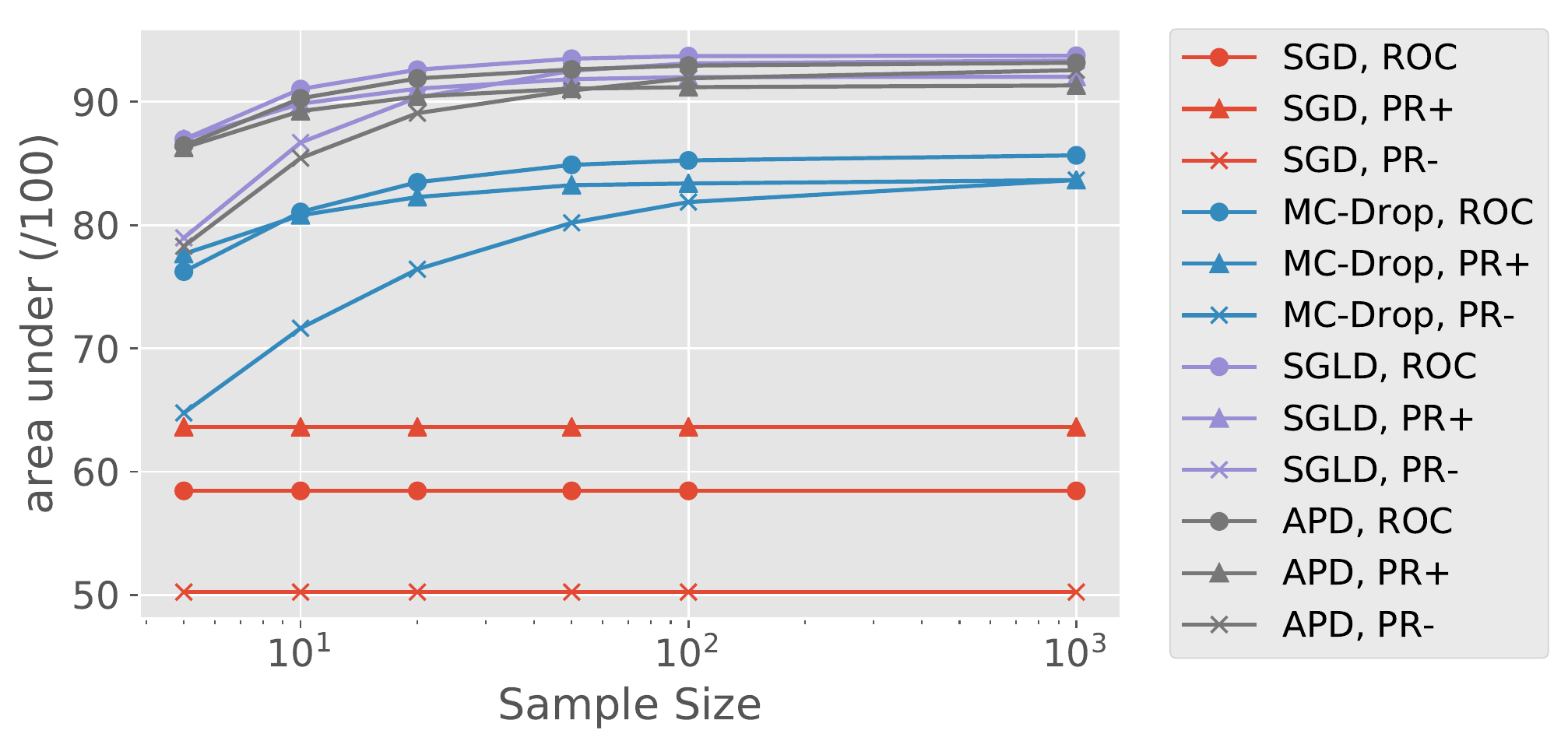}
    \vspace{-0.4cm}
    \caption{\textbf{Effect of Sample Size on Anomaly Detection.} We see improved performance with increasing sample size for test-time inference (10,000 total samples). Here, we measured uncertainty using VR.}
    \label{fig:anon-sample-size}
\end{figure}

We also investigated the impact of the sample size for test-time inference, using the \textbf{fcNN1} network and the notMNIST OOD dataset.
Figure~\ref{fig:anon-sample-size} shows that as we increased the sample size, anomaly detection performance improved.
Both SGLD and APD consistently outperformed MC dropout.

\begin{table*}[]
\centering
\scriptsize
\setlength\tabcolsep{3.5pt}
\begin{tabular}{@{}llcccccccccccc@{}}
\toprule
&\multicolumn{1}{c}{\textbf{Dataset}} & \multicolumn{3}{c}{\textbf{SGD}}           & \multicolumn{3}{c}{\textbf{MC-Dropout}}    & \multicolumn{3}{c}{\textbf{SGLD}}          & \multicolumn{3}{c}{\textbf{APD (Ours)}}    \\ \midrule
\textbf{Det.}&\multicolumn{1}{c}{\textbf{area under}}        & \textbf{ROC} & \textbf{PR+} & \textbf{PR-} & \textbf{ROC} & \textbf{PR+} & \textbf{PR-} & \textbf{ROC} & \textbf{PR+} & \textbf{PR-} & \textbf{ROC} & \textbf{PR+} & \textbf{PR-}   \\ \midrule
\multirow{5}{*}{\textbf{VR}}&\textbf{notMNIST}   &  64.2 & 67.6 & 54.4   & 88.0 & 87.2 & 82.1   & 98.1 & 97.8 & 98.3   & 97.8 & 97.4 & 98.1    \\
&\textbf{OmniGlot}                               &  84.2 & 84.9 & 78.7   & 91.5 & 90.8 & 90.3   & 99.0 & 98.8 & 99.1   & 98.8 & 98.6 & 99.1    \\
&\textbf{CIFAR10bw}                              &  61.4 & 66.1 & 52.2   & 90.1 & 88.5 & 86.5   & 97.4 & 97.0 & 97.5   & 96.9 & 96.5 & 96.7    \\
&\textbf{Gaussian}                               &  67.3 & 70.2 & 57.4   & 91.3 & 89.8 & 89.0   & 99.6 & 99.6 & 99.7   & 99.6 & 99.5 & 99.6    \\
&\textbf{Uniform}                                &  85.4 & 80.7 & 85.8   & 93.6 & 91.2 & 94.8   & 99.8 & 99.8 & 99.9   & 99.8 & 99.7 & 99.8    \\
\midrule
\multirow{5}{*}{\textbf{BALD}}&\textbf{notMNIST}    &  -  &  -  &    -   & 87.0 & 85.0 & 81.0    & 99.7  & 99.8  & 99.6     & 99.6  & 99.7  & 99.5     \\
&\textbf{OmniGlot}    &  -  &  -  &    -   & 91.4 & 90.7 & 90.5    & 99.9  & 100.0 & 99.9     & 99.9  & 99.9  & 99.9     \\
&\textbf{CIFAR10bw}   &  -  &  -  &    -   & 89.3 & 86.2 & 86.0    & 99.4  & 99.4  & 99.2     & 99.1  & 99.3  & 98.3     \\
&\textbf{Gaussian}    &  -  &  -  &    -   & 90.9 & 88.6 & 89.3    & 100.0 & 100.0 & 100.0    & 100.0 & 100.0 & 100.0    \\
&\textbf{Uniform}     &  -  &  -  &    -   & 97.3 & 96.6 & 97.9    & 100.0 & 100.0 & 100.0    & 100.0 & 100.0 & 100.0    \\
\bottomrule
\end{tabular}
\caption{\textbf{MNIST Anomaly Detection Results with fcNN2 (784-400-400-10).} We use variations ratio (top), and BALD (bottom). We show anomaly detection results on several OOD datasets \citep{hendrycks2016baseline}, with OOD data scaled by a factor of 5.}
\label{mnistanomalydetection}
\vspace{-.5em}
\end{table*}

\vspace{-0.1cm}
\subsection{Active Learning}

\begin{figure}[t]
    \centering
    \includegraphics[width=0.9
    \linewidth]{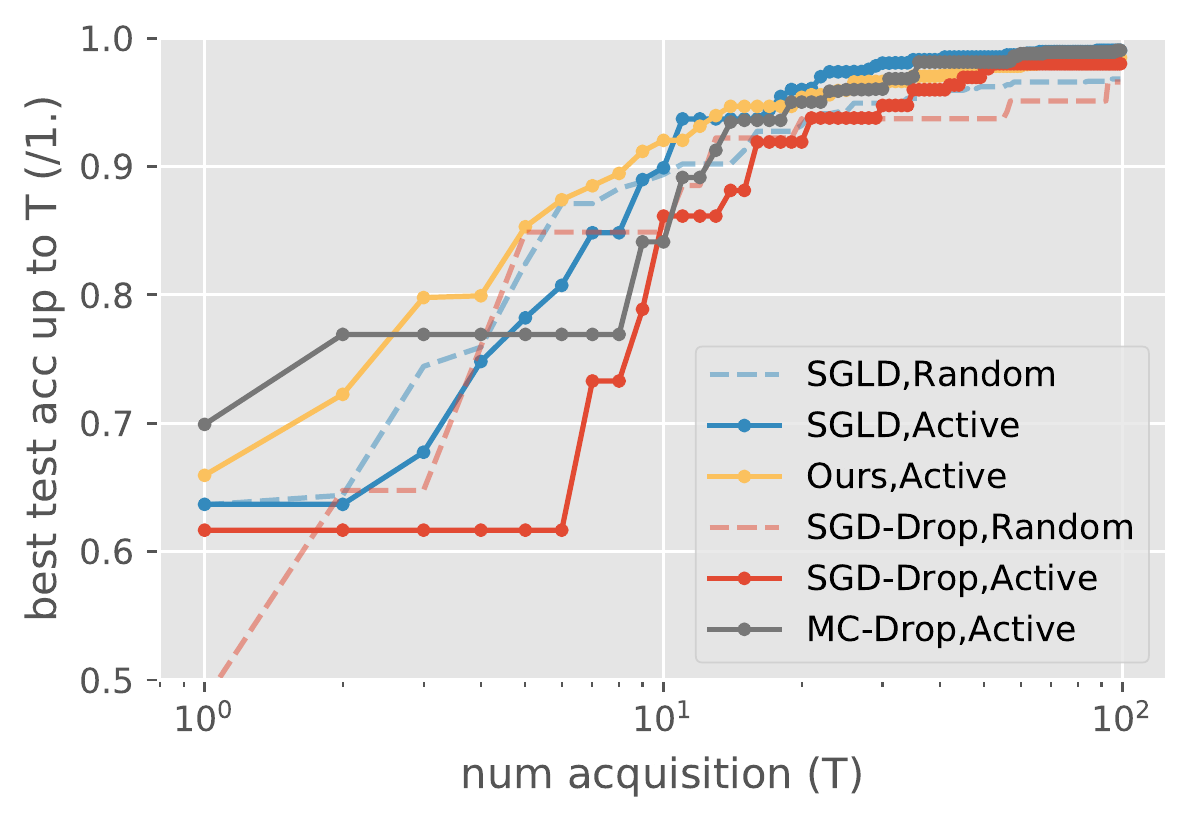}
    \caption{\textbf{Active Learning Results.} For BNNs, active learning is able to learn faster than random acquisition. SGD-Drop is SGD with dropout during training.}
    \label{fig:act-cold}
    \vspace{-0.4cm}
\end{figure}

For active learning, we followed the experimental setup of \citet{gal2017deep}, and evaluated on MNIST. For the acquisition function we used entropy, which performed well in \citet{gal2017deep}. The architecture for our prediction model was based on that of \citet{springenberg2014striving}. Instead of ReLU, we used LeakyReLU with a negative slope of 0.2, and we used half the number of filters of the original architecture.
Our initial training set consisted of 20 labeled images, 2 from each of the 10 digit classes; the rest of the images formed a \textit{poolset}. In each acquisition iteration, the model used an acquisition function to choose a set of 10 images from the pool to be labeled (e.g., by a human or oracle). Figure \ref{fig:act-cold} shows that BNNs outperform point-estimate counterparts consistently, and that using entropy as the acquisition function is better than random.
Our method performs best early (i.e., $<10$ acquisitions) due to either a better regularization effect or better uncertainty for active learning.

\vspace{-0.1cm}
\subsection{Adversarial Example Detection}

\begin{table}[]
\centering
\small
\begin{tabular}{@{}c|cccc@{}}
\toprule
\textbf{Source} & \textbf{Attack Type} & \textbf{MC-Drop} & \textbf{SGLD} & \textbf{Ours} \\ \hline
\multirow{2}{*}{\textbf{MC-Drop}} & \textbf{FGSM} & 89.53 & \textbf{94.01} & 91.70 \\
                                     & \textbf{PGD}  & 88.37 & \textbf{93.95} & 91.63 \\ \hline
\multirow{2}{*}{\textbf{SGLD}} & \textbf{FGSM} & 54.99 & \textbf{83.76}  & 75.93 \\
                               & \textbf{PGD}  & 56.91 & \textbf{84.98} & 82.80 \\ \hline
\multirow{2}{*}{\textbf{Ours}} & \textbf{FGSM} & 54.51 & 83.05 & \textbf{86.02} \\
                               & \textbf{PGD}  & 54.98 & 88.01 & \textbf{93.15} \\ \bottomrule
\end{tabular}
\caption{\textbf{MNIST Adversarial Detection Results.} Each row shows the AUROC for FGSM and PGD adversaries under each source model.}
\label{table:advdetection}
\vspace{-0.5cm}
\end{table}

Adversarial examples are inputs to a classifier which have been maliciously designed to force misclassification. These inputs are typically produced by taking some existing data point and applying a small perturbation to cause inaccurate predictions \cite{szegedy2013intriguing, goodfellow2014explaining}. Given an input data point $\bx$ which is correctly classified as $y$, a small perturbation, $\bdelta$, is added such that $\hat{\bx} = \bx + \bdelta$ is now classified as $\hat{y} \neq y$ by the classifier. Different choices of attack determine the form of the perturbation. In practice, we can often find $\bdelta$ which is imperceptibly small.

We focused on two attacks: Fast Gradient-Sign Method (FGSM) \cite{goodfellow2014explaining}, and Projected Gradient Descent (PGD) \cite{kurakin2016adversarial, madry2017towards}. FGSM utilizes a simple unit step in the direction of increasing gradient of the network's cost function. PGD utilizes repeated smaller steps of the same form, while projecting the output onto $B_\epsilon(\bx)$, the $\ell$-infinity ball of size $\epsilon$. FGSM is considered a relatively easy attack to defend against, while PGD is a strong attack with evidence supporting it as a ``universal'' first-order attack \cite{madry2017towards}. We made use of the foolbox library \cite{rauber2017foolbox} for both of these attacks, using our own implementation for PGD.

In this work, the adversary has access to the network architecture and a single posterior sample. We generated 6000 adversarial examples from the validation set using this posterior sample as fixed network weights. We then used 1000 samples from the posterior distribution to detect whether an input point was an adversarial example or belonged to the test set. We refer to this scheme as a gray-box attack because the attacker does not have access to the full posterior distribution. For these experiments, we used the approximate model variance (Eqn.~\ref{eqn:uncert}) which we found outperformed entropy and BALD on this problem.

Table \ref{table:advdetection} shows the results of using various MC techniques to detect adversarial examples.
We compared SGLD and our method to MC dropout inference detection \cite{feinman2017detecting} as a baseline. We used a small single-hidden-layer network with 100 units. For each detection method, we generated attacks using samples from the source model and tested these attacks against all other approaches.

We found that all three approaches were effective when detecting adversarial examples crafted with their own networks. However, when transferring attacks between networks there was a steep drop-off for MC dropout inference which performed only slightly better than random. Both SGLD inference and our own method were able to detect transferred attacks. In this setting we were able to see a more substantial gap between SGLD and our own method---suggesting that the GAN was unable to capture some quality of the posterior distribution that is critical for adversarial example detection.

Finally, we note that previous attempts at detection have proven ill-tested. Following the guidance of \citet{carlini2017adversarial}, we argue that it is critical to evaluate these methods on a more challenging dataset such as CIFAR-10 \cite{krizhevsky2009learning} and using attacks which take advantage of the detection scheme. We hope to explore this in future work.
\vspace{-0.1cm}
\subsection{Distillation with GANs}
Here, we show that distilling the parameters of a network is not trivial, and that recent advances in GANs make them a promising approach.
For the experiments in this section, we used the \textbf{fcNN1} architecture and measured performance on the anomaly detection task (see Section \ref{sec:anomdet}) with the notMNIST OOD dataset (as it is one of the most challenging ones).

\vspace{-0.4cm}
\paragraph{Do we need a GAN?} Using MCMC methods allows us to avoid making simplifying assumptions about the posterior distribution. In this section we show evidence that the multimodal posterior distribution induced by SGLD samples cannot be completely represented with simple, fully factorized approximations. We performed this analysis using a series of Mixture of Gaussians (MoG) with increasing number of components, $N_c$. We fit the MoG to posterior samples using the EM algorithm \cite{dempster1977maximum}.

\begin{figure}[htbp]
    \centering
    \vspace{-0.3cm}
    \includegraphics[width=0.85\linewidth]{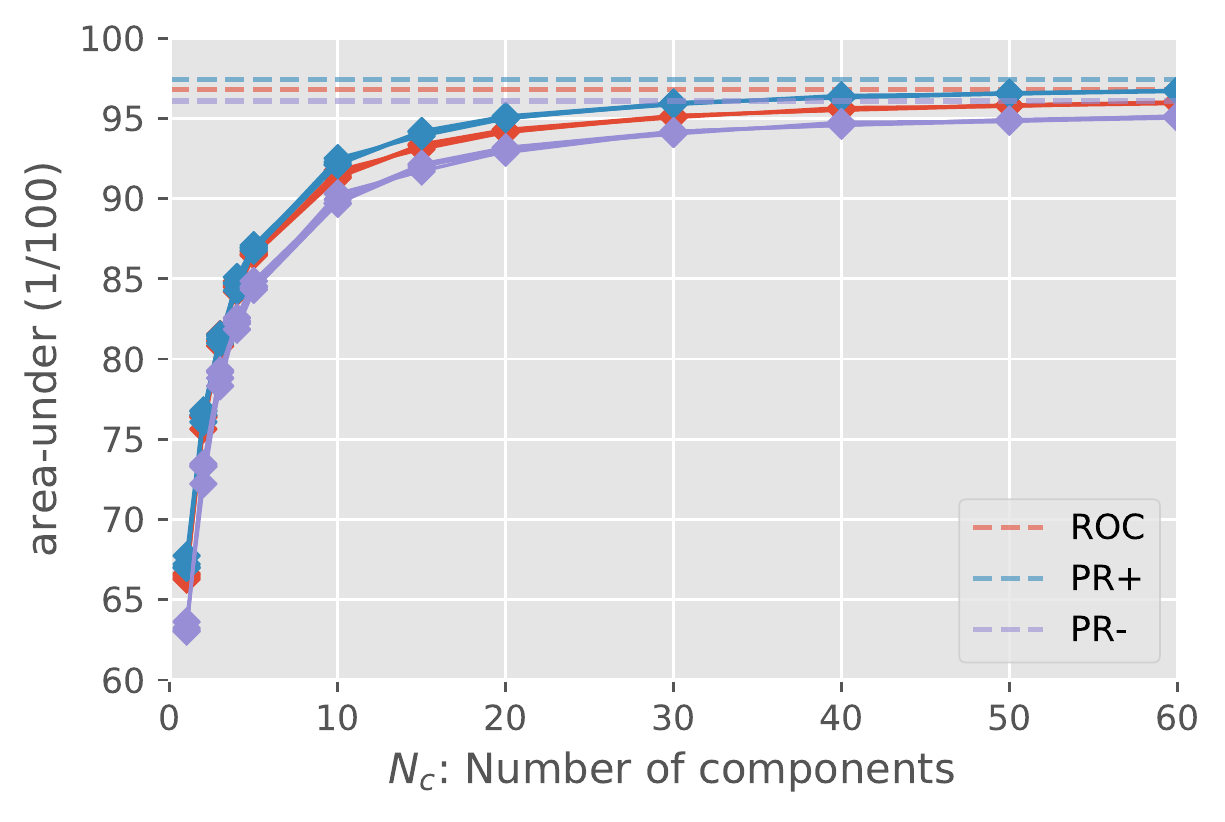}
    \caption{\textbf{Effect of Increasing Number of GMM Components on Anomaly Detection.} The dashed lines correspond to the performance when using SGLD samples used to train the GMM directly. We see improved performance as the number of components increases---approaching the performance of using the SGLD samples directly.}
    \label{fig:gmm_detection}
    \vspace{-0.2cm}
\end{figure}

We compared the performance of MoG with varying $N_c$ for anomaly detection in Figure \ref{fig:gmm_detection}. For each value of $N_c$ the MoG was trained using 2000 posterior samples drawn using SGLD, and the uncertainty measure was entropy. The performance of using SGLD directly is plotted as a dashed line. We see that when using a single component (corresponding to a fully factorized Gaussian) the performance is poor---suggesting that a single mode is not sufficient to model the posterior. As the number of components increases, the MoG begins to approach the performance achieved by using the SGLD samples themselves.

With enough components, the MoG is able to perform well on the anomaly detection tasks. In order to do so, the diagonal covariance MoG requires at least a mean and variance parameter for each network parameter per component. This reduces the memory overhead of using SGLD directly, but is still costly compared to the GAN.
Using a MoG model with 60 components (9.54M parameters) retains 99.3\% of the performance on this task w.r.t. the original SGLD samples.  APD (using WGAN-GP) retains 99.8\% of the performance while using fewer parameters (1.67M parameters with GAN hidden size 20).

 \begin{figure}[]
    \centering
    \vspace{-0.cm}
    \includegraphics[width=0.85\linewidth]{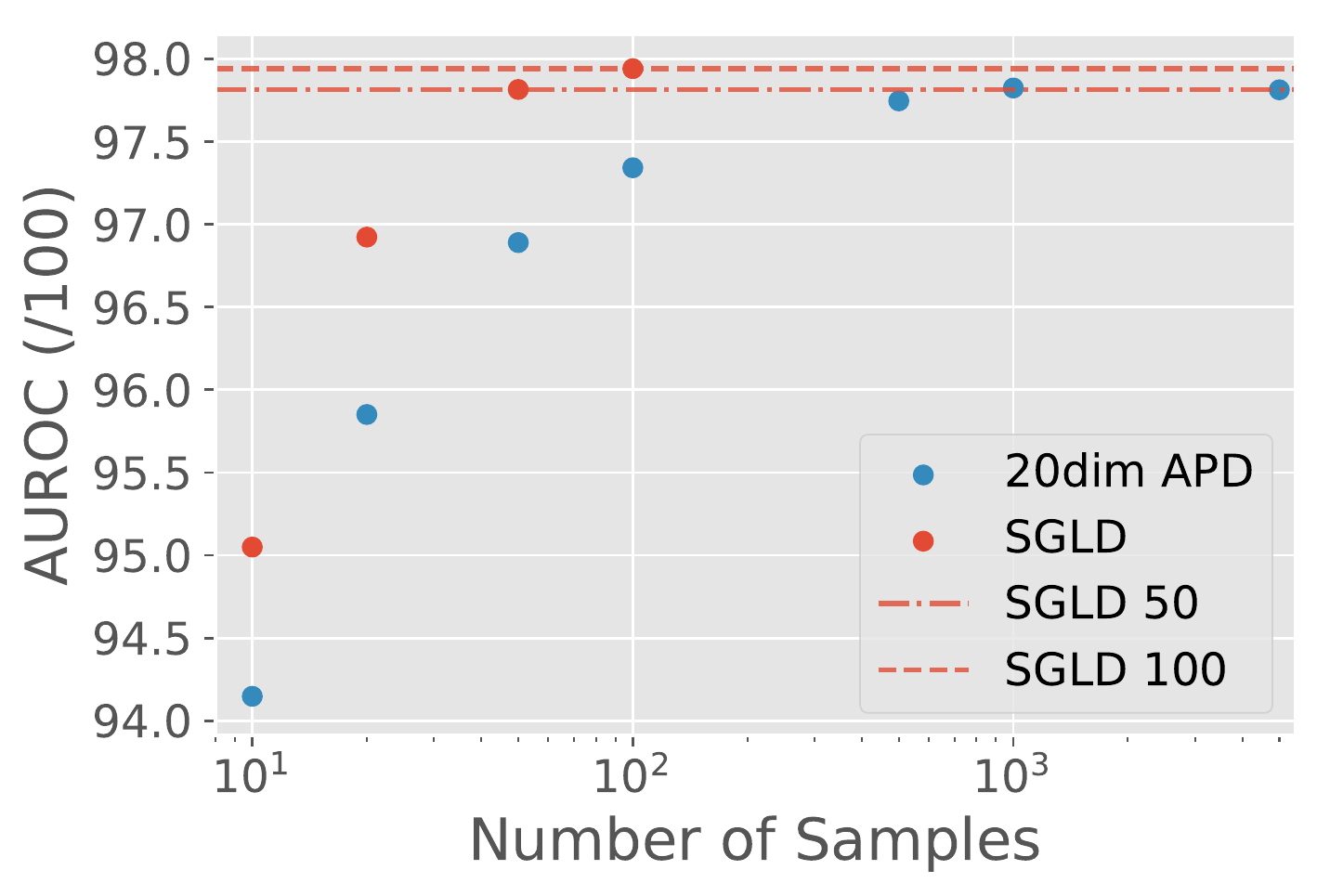}
    \vspace{-0.3cm}
    \caption
    {\textbf{Effect of SGLD and APD Sample Size.} With APD, the storage cost (i.e., generator size) is fixed regardless of the number of samples we generate at test-time. The two horizontal lines extend the y-value of SGLD at 50 and 100 samples. Here, we used BALD; VR and entropy yield similar results.}
    \label{fig:ss}
\end{figure}

\vspace{-0.2cm}
\paragraph{APD Storage Savings.}
We compared the performance of SGLD and APD using different numbers of samples at test-time.
With a 3-layer GAN with 20 hidden units per layer, generating 20 samples performs worse than simply using 20 original SGLD samples.
However, the storage cost of APD is not affected by drawing more samples. 
Figure~\ref{fig:ss} shows that as we generate more samples from the GAN, the performance improves and reaches that of 50 SGLD samples (i.e., ~2.5x storage savings).

\begin{figure}[]
 	\vspace{2mm}
    \centering
    \includegraphics[width=0.9\linewidth]{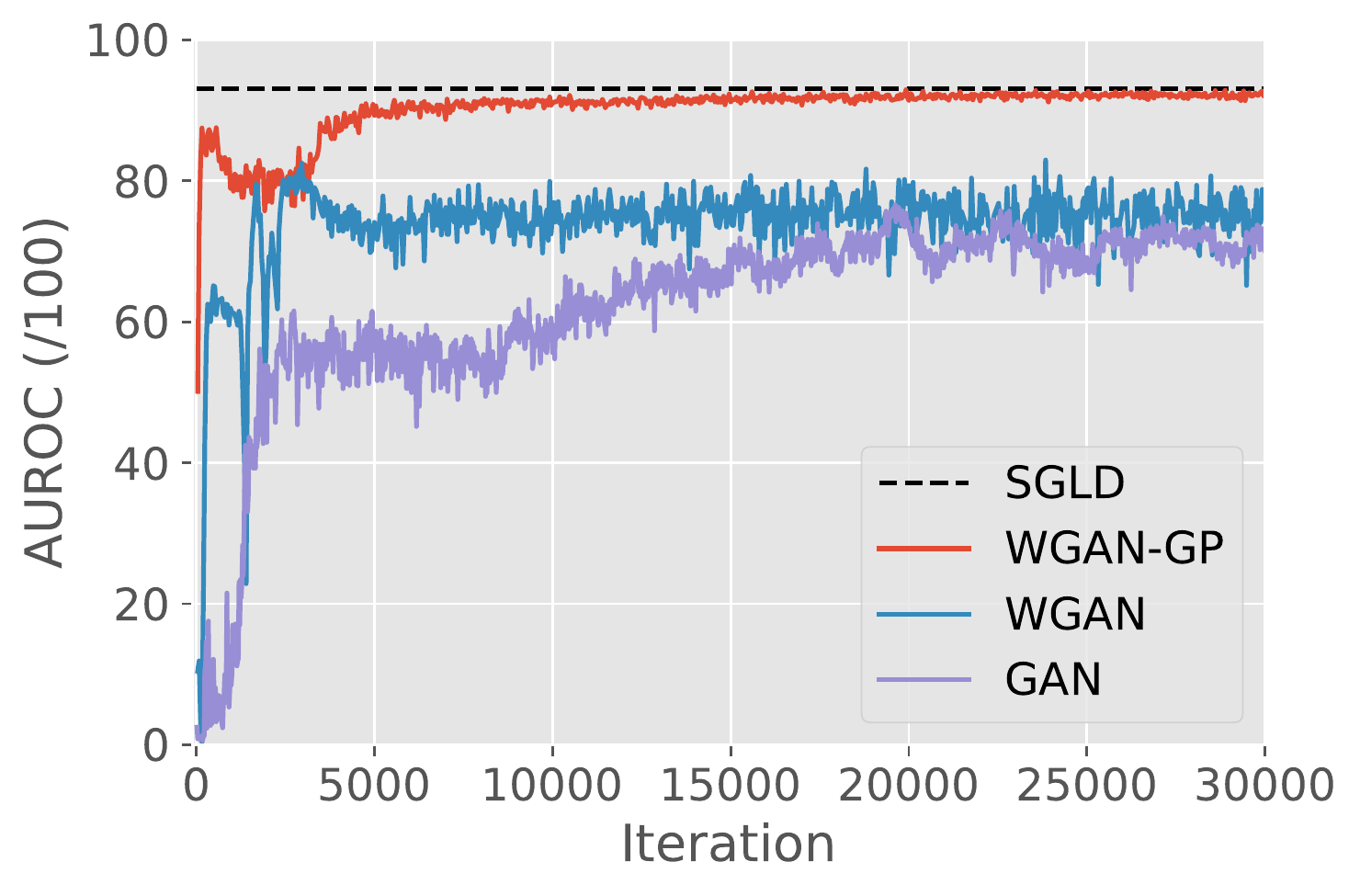}
    \vspace{-0.2cm}
    \caption
    {\textbf{Comparison of GAN Formulations.} WGAN-GP achieves better performance and converges faster than WGAN or the original GAN w.r.t. AUROC. Here, we used VR.}
    \label{fig:gancomparison}
    \vspace{-0.1cm}
\end{figure}

\vspace{-0.1cm}
\paragraph{Comparing GAN Formulations.}
Our framework makes use of recent advances in GANs.
We compared the training progress of our GAN using three popular variants: 1) the original formulation; 2) the Wasserstein GAN with weight-clipping; and 3) WGAN with gradient penalty.
Figure~\ref{fig:gancomparison} shows that WGAN-GP converges faster, and exhibits fewer oscillations from iteration to iteration.

\vspace{-0.1cm}
\section{Conclusion}
\label{sec_conclusion}

We introduced a framework for distilling BNN posterior samples drawn using SGLD, which we call Adversarial Posterior Distillation (APD).
Experimental results show that APD is able to retain the characteristics of the SGLD samples, as measured by performance on downstream applications including anomaly detection, active learning, and defense against adversarial attacks.
MCMC methods have attracted relatively little attention in BNNs due to their computational cost.
APD provides a way to reduce the storage cost. Our findings thus demonstrate that these MCMC methods have the potential to outperform simple alternatives such as MC dropout on important tasks that require uncertainty estimates. For future directions, we aim to explore other generative models that can further reduce the storage cost, such as autoregressive models.

\clearpage
\bibliography{asgld}
\bibliographystyle{icml2018}

\clearpage
\appendix

\section{Effect of Input Scale on Anomaly Detection}
\label{sec_A}
\begin{figure}[htbp]
    \centering
    \includegraphics[width=\linewidth]{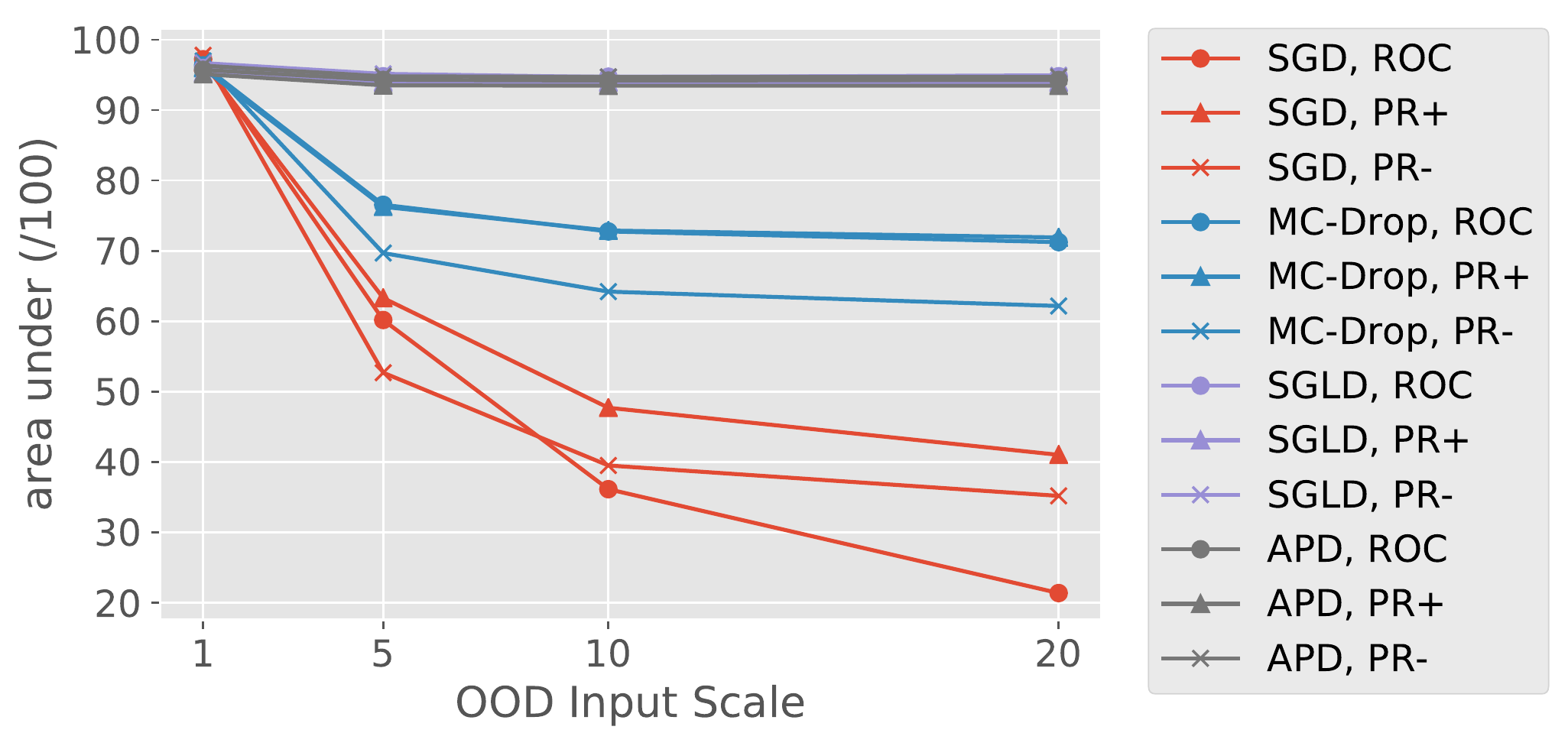}
    \caption{The effect of OOD pixel intensity scaling on difficulty of anomaly detection using the $\texttt{sklearn.datasets.load\_digits}$ dataset, with downsized notMNIST as the OOD data. Here, we used a small fully-connected network with architecture 64-100-10, and measured uncertainty using variation ratios.}
    \label{fig:ood-scale-baby}
\end{figure}

\begin{figure}[htbp]
    \centering
    \includegraphics[width=\linewidth]{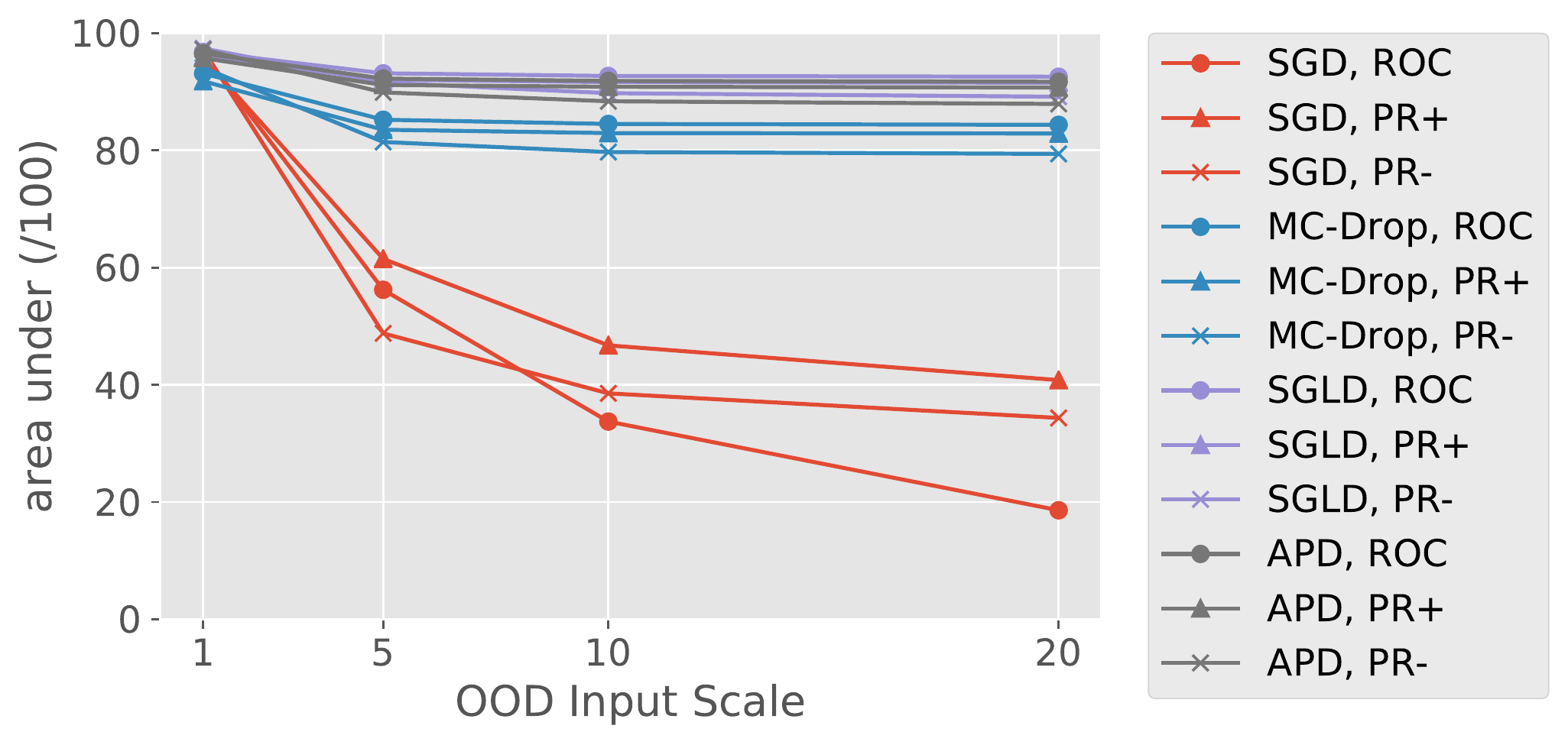}
    \caption{The effect of OOD pixel intensity scaling on difficulty of anomaly detection using MNIST, with notMNIST dataset OOD data. We used \textbf{fcNN1} and measured uncertainty using variation ratios.}
    \label{fig:ood-scale-mnist}
\end{figure}

In the anomaly detection benchmark introduced in \citet{hendrycks2016baseline}, the baseline models already perform very well; hence, the room to improve for BNNs is not very significant.
We performed an exploratory study in which we found that the baseline NN is highly susceptible to scaling of the pixel intensities of the out-of-distribution (OOD) data.
We performed this analysis on MNIST (using the notMNIST OOD dataset) as well as a smaller digit dataset, $\texttt{sklearn.datasets.load\_digits}$ (using downsized notMNIST as OOD data).
Figures~\ref{fig:ood-scale-baby} and~\ref{fig:ood-scale-mnist} illustrate the effect of OOD intensity scaling: the performance of the deterministic NN decreases as we increase the scaling factor; MC dropout is more resistant to increasing scale, while SGLD and APD are least affected.

\section{Additional Anomaly Detection Results}

We present comprehensive results for the anomaly detection task, with both \textbf{fcNN1} and \textbf{fcNN2} networks, and each of the variation ratios, entropy, and BALD uncertainty measures.
Table~\ref{table:mnistanomalydetection-fcnn1} shows the performance of SGD, MC dropout, SGLD, and APD using the \textbf{fcNN1} network (784-100-10) with VR, entropy, and BALD.
Table~\ref{table:mnistanomalydetection-fcnn2-entropy} shows the results of each method using the \textbf{fcNN2} network (784-400-400-10) with entropy (the results for \textbf{fcNN2} using VR and BALD are shown in Table~\ref{mnistanomalydetection} in Section~\ref{sec:anomdet}).

\begin{table*}[]
\centering
\footnotesize
\setlength\tabcolsep{3.5pt}
\begin{tabular}{@{}llcccccccccccc@{}}
\toprule
&\multicolumn{1}{c}{\textbf{Dataset}} & \multicolumn{3}{c}{\textbf{SGD}}           & \multicolumn{3}{c}{\textbf{MC-Dropout}}    & \multicolumn{3}{c}{\textbf{SGLD}}          & \multicolumn{3}{c}{\textbf{APD (Ours)}}    \\ \midrule
\textbf{Det.}&\multicolumn{1}{c}{\textbf{area under}}        & \textbf{ROC} & \textbf{PR+} & \textbf{PR-} & \textbf{ROC} & \textbf{PR+} & \textbf{PR-} & \textbf{ROC} & \textbf{PR+} & \textbf{PR-} & \textbf{ROC} & \textbf{PR+} & \textbf{PR-}   \\ \midrule
\multirow{5}{*}{\textbf{VR}}&\textbf{notMNIST}  & 58.5 & 63.6 & 50.2 & 86.5 & 84.4 & 84.9   & 93.2 & 92.0 & 92.1 & 92.7 & 91.7 & 91.1  \\
&\textbf{OmniGlot}  & 81.8 & 83.2 & 75.3 & 91.0 & 89.9 & 91.2   & 96.6 & 95.9 & 97.2 & 96.4 & 95.7 & 97.0  \\
&\textbf{CIFAR10bw} & 58.5 & 63.8 & 50.2 & 87.2 & 85.4 & 85.0   & 93.3 & 92.1 & 91.5 & 93.0 & 93.0 & 90.8  \\
&\textbf{Gaussian}  & 62.6 & 66.2 & 53.5 & 90.4 & 88.6 & 90.1   & 99.1 & 98.8 & 99.3 & 99.0 & 98.8 & 99.2  \\
&\textbf{Uniform}   & 91.7 & 89.8 & 92.6 & 91.8 & 89.5 & 93.2   & 99.4 & 99.2 & 99.5 & 99.3 & 99.1 & 99.4  \\
\midrule
\multirow{5}{*}{\textbf{Entropy}}&\textbf{notMNIST}   &  57.7 & 61.4 & 49.9   & 83.3 & 80.2 & 79.7   & 92.3 & 90.9 & 91.1   & 92.2 & 91.0 & 90.7    \\
&\textbf{OmniGlot}   &  83.5 & 84.4 & 78.5   & 91.0 & 90.2 & 91.1   & 96.4 & 95.9 & 96.7   & 96.7 & 96.3 & 97.0    \\
&\textbf{CIFAR10bw}  &  56.8 & 61.2 & 49.3   & 85.6 & 82.2 & 83.6   & 91.1 & 89.9 & 87.8   & 92.3 & 90.9 & 90.5    \\
&\textbf{Gaussian}   &  61.3 & 63.5 & 52.7   & 89.6 & 87.2 & 89.1   & 99.2 & 99.0 & 99.3   & 99.3 & 99.2 & 99.4    \\
&\textbf{Uniform}    &  95.1 & 94.0 & 95.9   & 94.8 & 93.2 & 95.7   & 99.7 & 99.6 & 99.8   & 99.8 & 99.7 & 99.8    \\
\midrule
\multirow{5}{*}{\textbf{BALD}}&\textbf{notMNIST}   &  -  &  -  &  -  & 83.4 & 80.4 & 79.6   & 97.8  & 98.3  & 95.8    & 97.6  & 98.3  & 95.5    \\
&\textbf{OmniGlot}   &  -  &  -  &  -  & 91.0 & 90.2 & 91.2   & 99.1  & 99.2  & 99.0    & 99.2  & 99.3  & 99.1    \\
&\textbf{CIFAR10bw}  &  -  &  -  &  -  & 85.7 & 82.4 & 83.6   & 96.8  & 97.7  & 93.2    & 97.6  & 98.2  & 94.8    \\
&\textbf{Gaussian}   &  -  &  -  &  -  & 89.6 & 87.2 & 89.1   & 100.0 & 100.0 & 100.0   & 100.0 & 100.0 & 100.0   \\
&\textbf{Uniform}    &  -  &  -  &  -  & 94.7 & 93.1 & 95.6   & 100.0 & 100.0 & 100.0   & 100.0 & 100.0 & 100.0   \\
\bottomrule
\end{tabular}
\caption{\textbf{MNIST Anomaly Detection Results with fcNN1 (784-100-10) using VR, entropy, and BALD uncertainty measures.}}
\label{table:mnistanomalydetection-fcnn1}
\end{table*}

\begin{table*}[]
\centering
\footnotesize
\setlength\tabcolsep{3.5pt}
\begin{tabular}{@{}llcccccccccccc@{}}
\toprule
\multicolumn{1}{c}{\textbf{Dataset}} & \multicolumn{3}{c}{\textbf{SGD}}           & \multicolumn{3}{c}{\textbf{MC-Dropout}}    & \multicolumn{3}{c}{\textbf{SGLD}}          & \multicolumn{3}{c}{\textbf{APD (Ours)}}    \\ \midrule
\textbf{Det.}&\multicolumn{1}{c}{\textbf{area under}}        & \textbf{ROC} & \textbf{PR+} & \textbf{PR-} & \textbf{ROC} & \textbf{PR+} & \textbf{PR-} & \textbf{ROC} & \textbf{PR+} & \textbf{PR-} & \textbf{ROC} & \textbf{PR+} & \textbf{PR-}   \\ \midrule
\multirow{5}{*}{\textbf{Entropy}}&\textbf{notMNIST}   &  61.4 & 64.4 & 52.7   & 87.0 & 85.0 & 81.0   & 98.7 & 98.5 & 98.8   & 98.3 & 98.0 & 98.4    \\
&\textbf{OmniGlot}   &  84.0 & 85.2 & 78.7   & 91.4 & 90.7 & 90.6   & 99.4 & 99.4 & 99.5   & 99.3 & 99.2 & 99.4    \\
&\textbf{CIFAR10bw}  &  59.9 & 65.0 & 51.2   & 89.1 & 86.2 & 85.4   & 97.8 & 97.5 & 97.9   & 97.2 & 96.8 & 96.8    \\
&\textbf{Gaussian}   &  64.2 & 67.4 & 54.7   & 90.9 & 88.6 & 89.5   & 99.8 & 99.8 & 99.9   & 99.8 & 99.7 & 99.8    \\
&\textbf{Uniform}    &  86.4 & 83.0 & 86.1   & 97.3 & 96.5 & 97.8   & 99.9 & 99.8 & 99.9   & 99.9 & 99.8 & 99.9    \\
\bottomrule

\end{tabular}
\caption{\textbf{MNIST Anomaly Detection Results with fcNN2 using entropy.} (The results using VR and BALD are given in the main paper.)}
\label{table:mnistanomalydetection-fcnn2-entropy}
\end{table*}

\end{document}